\documentclass[11pt]{article}

\usepackage{authblk}
\usepackage[a4paper]{geometry}
\usepackage{clic2017}
\usepackage{times}
\usepackage[utf8]{inputenc}
\usepackage[pdftex,bookmarks=true]{hyperref}
\hypersetup{colorlinks = true,  linkcolor = red,   anchorcolor = red,    citecolor = blue,  filecolor = red,   urlcolor = red}

\usepackage{url}
\usepackage{latexsym}
\usepackage{bm}
\usepackage{booktabs}
\usepackage{lipsum}
\usepackage{graphicx}

\usepackage{pgfplots}
\usepackage{caption}
\usepackage{subcaption}
\usepackage{standalone}
\usepackage{todonotes}
\usepackage{comment}

\usepackage{amsmath}

\DeclareMathOperator*{\argmin}{argmin} %

\renewcommand{\vec}[1]{\mathbf{#1}}

\title{UWB @ DIACR-Ita: Lexical Semantic Change Detection with CCA and Orthogonal Transformation}

\newcommand\blfootnote[1]{%
  \begingroup
  \renewcommand\thefootnote{}\footnote{#1}%
  \addtocounter{footnote}{-1}%
  \endgroup
}

	\author[* 1,2]{\bf Ond\v{r}ej Pra\v{z}\'{a}k}
	\author[* 1,2]{\bf Pavel P\v{r}ib\'{a}\v{n}}
	\author[* 2]{\bf  Stephen Taylor}
	\affil[1]{NTIS -- New Technologies for the Information Society,}
	\affil[2]{Department of Computer Science and Engineering,}
	\affil[ ]{Faculty of Applied Sciences, University of West Bohemia, Czech Republic}
	\affil[  ]{\tt	\{ondfa, pribanp, taylor\}@kiv.zcu.cz}
	\affil[  ]{\tt http://nlp.kiv.zcu.cz}

\date{}

\begin{document}
\maketitle
\begin{abstract}
  In this paper, we describe our method for detection of lexical semantic change (i.e., word sense changes over time) for the DIACR-Ita shared task, where we ranked $1^{st}$. We examine semantic differences between specific words in two Italian corpora, chosen from different time periods.
Our method is fully unsupervised and language independent. It consists of preparing a semantic vector space for each corpus, earlier and later. Then we compute a linear transformation between earlier and later spaces, using CCA and Orthogonal Transformation. Finally, we measure the cosines between the transformed vectors.

\end{abstract}
\blfootnote{
    \hspace{-0.65cm}  
    ${}^{\text{*}}$Equal contribution. Copyright \textcopyright\  
2020 for this paper by its authors. Use permitted under Creative Commons License Attribution 4.0 International (CC BY 4.0).
}

\vspace{-0.9cm}
\section{Introduction}
Language evolves with time. New words appear, old words fall out of use, and the meanings of some words shift.
There are changes in topics, syntax, and presentation structure. Reading the natural philosophy musings of aristocratic amateurs from the
eighteenth century, and comparing with a monograph from the nineteenth century, or a medical study from the twentieth century, we can observe differences in many dimensions, some of which need a deep historical background to study. Changes in word senses are both a visible and a tractable part of language evolution. 

Computational methods for researching the stories of words  have the potential of helping us understand this small corner of linguistic evolution. The tools for measuring these diachronic semantic shifts might also be useful for measuring whether the same word is  used in different ways in synchronic documents. The task of finding word sense changes over time is called diachronic \textit{Lexical Semantic Change (LSC)} detection. The task
is getting more attention in recent years \cite{hamilton-etal-2016-diachronic,schlechtweg-etal-2017-german,sem20-task1-overview}.
There is also the \textit{synchronic} \textit{LSC} task, which aims to identify domain-specific changes of word senses compared to general-language usage \cite{schlechtweg-etal-2019-wind}.

\subsection{Related Work}
\newcite{tahmasebi2018survey} provide a comprehensive survey of  techniques for the \textit{LSC} task, as do \newcite{kutuzov-etal-2018-diachronic}. \newcite{schlechtweg-etal-2019-wind} evaluate available approaches for \textit{LSC} detection  using the \textit{DURel} dataset \cite{schlectweg-etal-DURel}. \newcite{sem20-task1-overview} present results of the first shared task that addresses the LSC problem and provide an evaluation dataset that was manually annotated for four languages.
 
According to \newcite{schlechtweg-etal-2019-wind}, there are three  main types of approaches. (1) Semantic vector spaces approaches \cite{gulordava-baroni-2011-distributional,eger-mehler-2016-linearity,hamilton-etal-2016-cultural,hamilton-etal-2016-diachronic,rosenfeld-erk-2018-deep,uwb-semeval-2020} represent each word with two vectors for two different time periods. The change of meaning is then measured by some distance (usually by the cosine distance)
between the two vectors. (2) Topic modeling approaches 
\cite{bamman-topics,mihalcea-nastase-2012-word,cook-etal-2014-novel,frermann-lapata-2016-bayesian,Schlechtweg20} estimate a probability distribution of words over their different senses, i.e., topics and (3) Clustering models \cite{mitra2015automatic,tahmasebi-risse-2017-finding}. 

\subsection{The DIACR-Ita task}
The goal of the DIACR-Ita task \cite{diacritaevalita2020} is to establish if a set of Italian words (target words) change their meaning from time period $t_1$ to time period $t_2$ (i.e., binary classification task).
The organizers provide corresponding corpora $C_1$ and $C_2$ and a list of target words.  Only these inputs may be used to train systems, which judge for each target word, whether it is changed or not. The task is the same as the binary sub-task of the SemEval-2020 Task 1 \cite{sem20-task1-overview} competition.


\section{Data}
The DIACR-Ita data consists of many randomly ordered text samples that have no relationship to each other.
Most of the text samples are complete sentences, but some are sentence fragments.

The `early' corpus, $C_1$  
has about 2.4 million text samples and 52 million tokens; the
`later' corpus, 
$C_2$ has about 7.8 million text samples and 738 million
tokens.
Each token is given in the corpora with its part-of-speech tag and lemma.
The target word list consists of 18 lemmas.  
The POS and lemmas of the corpora are generated with the
UDPipe \cite{straka-2018-udpipe} model ISDT-UD v2.5, which has an error rate of about 2\%.


\section{System Description}
\subsection{Overview}
Because language is evolving, 
expressions, words, and sentence constructions in two corpora from different time periods about the same topic
will be written in languages that are quite similar but slightly different. They will 
share the 
majority of their words, grammar, and syntax.
We can observe a similar situation in languages from the same family, such as \textit{Italian-Spanish} in Romance languages or \textit{Czech-Slovak} in Slavic languages. These pairs of languages share a lot of common words, expressions and syntax. For some pairs, native speakers can understand and sometimes even actively communicate through a (low) language barrier.

\par Our system follows the approach from \cite{uwb-semeval-2020}\footnote{The source code is available at \url{https://github.com/pauli31/SemEval2020-task1}}. The main idea behind our solution is that we treat each pair of corpora $C_1$ and $C_2$ as different languages $L_1$ and $L_2$ even though the text from both corpora is written in Italian. 
We believe that these two languages $L_1$ and $L_2$ will be extremely similar in all aspects, including semantic. We train a separate  semantic space for each corpus, and subsequently, we map these two spaces into one common cross-lingual space. We use methods for cross-lingual mapping \cite{Brychcin2019,artetxe-labaka-agirre:2016:EMNLP2016,artetxe-etal-2017,artetxe-etal-2018b,artetxe-etal-2018-robust} and thanks to the large similarity between $L_1$ and $L_2$ the quality of transformation should be high. We compute cosine similarity of the transformed word vectors to classify
 whether the target words changed their sense.

\subsection{Semantic Space Transformation}
\par First, we train two semantic spaces from corpus $C_1$ and $C_2$. We represent the semantic spaces by a matrix $\vec{X}^s$ (i.e., a source space $s$) and a matrix $\vec{X}^t$ (i.e., a target space $t$)\footnote{The source space $\vec{X}^s$ is created from the corpus $C_1$ and the target space $\vec{X}^t$  is created from the corpus $C_2$.} using word2vec Skip-gram with negative sampling \cite{Mikolov2013a}. We perform a cross-lingual mapping of the two vector spaces,
getting two matrices $\vec{\hat{X}}^s$ and $\vec{\hat{X}}^t$ projected into a shared space. We select two methods for the cross-lingual mapping \textit{Canonical Correlation Analysis (CCA)} using the implementation from \cite{Brychcin2019} and a modification of the \textit{Orthogonal Transformation} from \textit{VecMap} \cite{artetxe-etal-2018-robust}. Both of these methods are linear transformations. The transformations can be written as follows:
\begin{equation}
  \vec{\hat{X}}^s =   \vec{W}^{s \rightarrow t} \vec{X}^s
\end{equation}
where $\vec{W}^{s \rightarrow t}$ is a matrix that performs linear transformation from the source space $s$ (matrix $\vec{X}^s$) into a target space $t$ and $\vec{\hat{X}}^s$ is the source space transformed into the target space $t$ (the matrix $\vec{X}^t$ does not have to be transformed because $\vec{X}^t$ is already in the target space $t$ and $\vec{X}^t = \vec{\hat{X}}^t$).

Finally, in all transformation methods, for each word $w_i$ from the set of target words $T$, we select its corresponding vectors $\vec{v}_{w_i}^{s}$ and $\vec{v}_{w_i}^{t}$ from matrices $\vec{\hat{X}}^s$ and $\vec{\hat{X}}^t$, respectively ($\vec{v}_{w_i}^{s} \in \vec{\hat{X}}^s$ and $\vec{v}_{w_i}^{t} \in \vec{\hat{X}}^t$), and we compute cosine similarity between these two vectors. The cosine similarity is then used to generate a final classification output using different strategies, see Section \ref{subsec:binary-system} and \ref{subsec:ranking-strategy}.

\subsection{Canonical Correlation Analysis}
\par Generally, the CCA transformation transforms both spaces $\vec{X}^s$ and $\vec{X}^t$ into a third shared space  $o$ (where $\vec{X}^s\neq \vec{\hat{X}}^s$ and $\vec{X}^t\neq \vec{\hat{X}}^t$). Thus, CCA computes two transformation matrices $\vec{W}^{s \rightarrow o}$ for the source space and $\vec{W}^{t \rightarrow o}$ for the target space. The transformation matrices are computed by minimizing the negative correlation between the vectors $\vec{x}_{i}^{s} \in \vec{X}^s$ and $\vec{x}_{i}^{t} \in \vec{X}^t$ that are projected into the shared space $o$. The negative correlation is defined as follows:

\vspace{-0.7cm}
\begin{equation}
\begin{split}
    \argmin_{\vec{W}^{s \rightarrow o}, \vec{W}^{t \rightarrow o}} -\sum_{i=1}^{n} \rho (\vec{W}^{s \rightarrow o}\vec{x}_{i}^{s}, \vec{W}^{t \rightarrow o} \vec{x}_{i}^{t} )  =  \\
    -\sum_{i=1}^{n}  \frac{cov(\vec{W}^{s \rightarrow o}\vec{x}_{i}^{s}, \vec{W}^{t \rightarrow o} \vec{x}_{i}^{t})}{\sqrt{var(\vec{W}^{s \rightarrow o}\vec{x}_{i}^{s}) \times var(\vec{W}^{t \rightarrow o} \vec{x}_{i}^{t})}}
     \end{split}
\end{equation}
where $cov$ is the covariance, $var$ is the variance and $n$ is the number of vectors used for computing the transformation. In our implementation of CCA, the matrix $\vec{\hat{X}}^t$ is equal to the matrix $\vec{X}^t$ because it transforms only the source space $s$ (matrix $\vec{X}^s$) into the target space $t$ from the common shared space with a pseudo-inversion, and the target space does not change. The matrix $\vec{W}^{s \rightarrow t}$ for this transformation is then given by:

\begin{equation}
     \vec{W}^{s \rightarrow t} = \vec{W}^{s \rightarrow o} (\vec{W}^{t \rightarrow o})^{-1}
\end{equation}

The submissions that use CCA are referred to as \textbf{cca-bin} and \textbf{cca-ranking} in Table \ref{tab:results}. The \textbf{-bin} and \textbf{-ranking} parts refer to a strategy used for the final classification decision, see Section \ref{subsec:binary-system} and \ref{subsec:ranking-strategy}.

\subsection{Orthogonal Transformation}
\par In the case of the Orthogonal Transformation, the submission is referred to as \textbf{ort-bin}. We use Orthogonal Transformation with a supervised seed dictionary consisting of all words common to both semantic spaces. The transformation matrix $\vec{W}^{s \rightarrow t}$ is given by:
\begin{equation}
    \argmin_{\vec{W}^{s \rightarrow t}}{\sum_i^{|V|}{(\vec{W}^{s \rightarrow t} \vec{x}^s_i - \vec{x}^t_i)^2}}
\end{equation}
under the hard condition that $\vec{W}^{s \rightarrow t}$ needs to be orthogonal, where V is the vocabulary of correct word translations from source space $\vec{X}^s$ to target space $\vec{X}^t$ and $\vec{x}_{i}^{s} \in \vec{X}^s$ and $\vec{x}_{i}^{t} \in \vec{X}^t$. The reason for the orthogonality constraint is that linear transformation with an orthogonal matrix does not squeeze or re-scale the transformed space. It only rotates the space, thus it preserves most of the relationships of its elements (in our case, it is important that orthogonal transformation preserves angles between the words, so it preserves the cosine similarity).



\subsection{Binary Strategy}
\label{subsec:binary-system}
We use different strategies for the binary classification output, but all have in common that they use continuous scores. The continuous score for each target word is computed as the cosine similarity between the two vectors from the earlier and later corpus.

\par In the case of the \textit{binary strategy}, we assume a threshold $t$ for which the target words with a continuous score greater than $t$ changed meaning and words with the score lower than $t$ did not. We know that this assumption is generally wrong (because using the threshold, we introduce some error into the classification), but we still believe it holds for most cases and it is the best choice.
To estimate the threshold $t$, we used an approach called \textit{binary-threshold} (\textbf{cca-bin} and \textbf{ort-bin} in Table \ref{tab:results}). For each target word $w_i$ we compute cosine similarity of its vectors $\vec{v}_{w_i}^{s}$ and $\vec{v}_{w_i}^{t}$, then we average these similarities for all words. The resulting averaged\footnote{The \textbf{ort-bin} submission sets the threshold to be in the largest gap between the similarity values} value is used as the threshold.


\subsection{Ranking Strategy}
\label{subsec:ranking-strategy}
The \textit{ranking strategy} is the second approach for generating a classification output (the submission result \textbf{cca-ranking} in Table \ref{tab:results}).
It uses the mean rank of repeated runs of each embedding pair.
For each run, the target words are scored with
a cosine distance.  Then the distances for each embedding pair are sorted and a rank-order is assigned
to each target.
The rank-orders are averaged, to get a mean rank (and a  
standard deviation) for each target for each pair.  
Finally, ranks for all embedding pairs are averaged. The composite rank is used, along with an
estimate of the associated cosine distance and its corresponding angle,
to divide
the target list into changed and unchanged sets.  This does not work well; there are competing gaps in rank and distance estimates.

We use the number of embeddings, and not the total number of runs, to compute the standard error of the mean (which is standard deviation divided by the square root of samples).

\section{Experimental Setup}
\label{sec:exp-setup}
To obtain the semantic spaces, we employ Skip-gram with negative sampling \cite{Mikolov2013a}. For the final submission, we trained the semantic spaces with $100$ (the \textbf{ort-bin} submission) and $150$ (the \textbf{cca-bin} submission) dimensions for five iterations with five negative samples and window size set to five. Each word has to appear at least five times in the corpus to be used in the training. To train the semantic space, we used the lemmatized corpora. The dimensions $100$ and $150$ are selected based on our previous experiences with these methods \cite{uwb-semeval-2020}. Since we were able to submit four different submissions, we did not use the same dimension for both methods.

\par The \textbf{cca-ranking} submission uses the same settings and dimensions 100-105, 110-115, etc. up to 210-215, resulting in 72 different dimension sizes. It combines 40 runs on each of 72 embedding pairs, a total of 2880 runs.

\par For the \textbf{cca-bin} submission, we build the translation dictionary for the transformation of the two spaces by removing the target words from the intersection of their vocabularies. In the case of the \textbf{cca-ranking} submission, the dictionary in each run consists of up to 5000 randomly chosen common words for each semantic space.

\par The \textbf{random} submission represents output that was generated completely randomly.

\subsection{Corpus variants}
The organizers provided the corpora already tokenized in four different versions: original tokens; lemmatized tokens; original tokens with POS tag; lemmatized tokens with POS tag.
%
We experimented with each of these variants, although in the end, we used
results based only on lemmas.
\begin{figure}[ht]
\begin{centering}
\includegraphics[width=7.8cm]{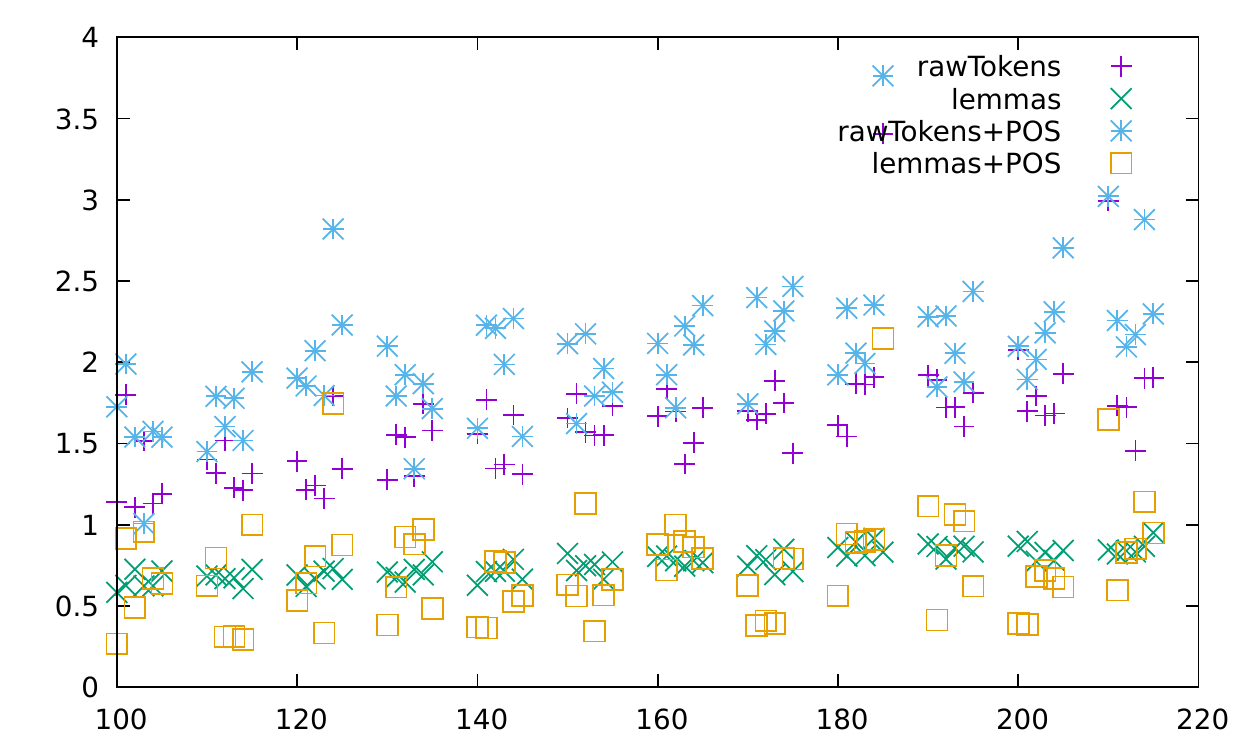}
\end{centering}
\caption{Standard deviation (of rank) versus embedding size for four versions
of the corpora.}
\label{fig:graph3}
\end{figure}
Figure \ref{fig:graph3} shows the mean standard deviation of rank for target words
over forty runs for each of 72 different embedding sizes. The most consistent variant is the \emph{lemmas} only.

\section{Results}
We submitted four different submissions. The accuracy results for each submission are shown in Table \ref{tab:results}. The \textbf{ort-bin} system achieved the best accuracy of $0.944$ and ranked first\footnote{We share the first place with another  team that achieved the same accuracy.} among eight other teams in the shared task,  classifying 17 out of 18 target words correctly. The \textbf{cca-bin} system achieved an accuracy of $0.889$ (16 correct classifications out of 18). After releasing the gold labels, we performed an additional experiment with the \textbf{cca-bin} system achieving also an accuracy of $0.944$ when the same word embeddings (with embeddings dimension 100 instead of 150) are used as for the \textbf{ort-bin} system. We found an optimal threshold for both systems, which makes them classify all the words correctly\footnote{That is, 100\% accuracy was possible with the continuous scores of both methods if we only had an oracle to set the threshold.}.

\par We believe that the key factor of the success of our system is the sufficient size of the provided corpora. Thanks to that, we were able to train semantic spaces of good quality and thus achieve good results.

\begin{table}[h!]
\centering
\begin{tabular}{lc}
\toprule
\textbf{System}      & \textbf{Accuracy} \\
\hline
cca-bin     & .889                        \\  
ort-bin     & .944                        \\  
cca-ranking & .778                        \\  
random      & .500   \\                     
\bottomrule
\end{tabular}
   \caption{Results for our final submissions.}
    \label{tab:results}
\end{table}

\newcommand\xmin{25}
\newcommand\ymin{0.42}
\newcommand\ymax{0.83}

\newcommand\xminrank{25}
\newcommand\yminrank{0.51}
\newcommand\ymaxrank{1.0}

\newcommand\heig{3.9cm}



\section{Conclusion}
Our systems based on Canonical Correlation Analysis and Orthogonal Transformation achieved the best accuracy of 0.944 in the shared task and ranked first among eight other teams. We showed that our approach is a suitable solution for the \textit{Lexical Semantic Change} detection task. 
Applying a threshold to semantic distance is a sensible architecture
for detecting the binary semantic change in target words between two corpora.  Our  \emph{binary-threshold} strategy succeeded quite well.  

This task provided plenty of text to build good word embeddings.  Corpora with much smaller amounts of data might have increased the random variation between the earlier and later embeddings, which would have given our method problems. 
A flaw in our technique is that semantic vectors are  based on all senses of a word in the corpus.  We do not yet have tools
to tease out what kinds of changes are implied by a particular semantic distance between vectors.
We considered using the part of speech data in the corpora since different parts of speech for the same lemma are likely different senses. But placing the POS in the
token, like using inflections instead of lemmas, results in many more, less well-trained semantic vectors, as suggested by Figure 1.



\section*{Acknowledgements}
This work has been partly supported by ERDF ”Research and Development of Intelligent Components of Advanced Technologies for the Pilsen Metropolitan Area (InteCom)” (no.:   CZ.02.1.01/0.0/0.0/17 048/0007267); by the project LO1506 of the Czech Ministry of Education, Youth and Sports; and by Grant No. SGS-2019-018 Processing of heterogeneous data and its specialized applications.  Access to computing and storage facilities owned by parties and projects contributing to the National Grid Infrastructure MetaCentrum provided under the programme "Projects of Large Research, Development, and Innovations Infrastructures" (CESNET LM2015042), is greatly appreciated.
\bibliography{sem20}
\bibliographystyle{acl.bst}

\end{document}